\documentclass[10pt, conference]{IEEEtran}
\usepackage{amsmath,amssymb,amsfonts}
\usepackage{caption}
\usepackage{algorithmic}
\usepackage{array}
\usepackage{graphicx}
\usepackage{textcomp}
\usepackage{xcolor}
\usepackage{url}
\usepackage{siunitx}
\usepackage{pgfplots}
\pgfplotsset{compat=newest}
\usepackage{booktabs}
\usepackage{multirow}
\usepackage{float}
\usepackage{relsize}
\usepackage{pdfpages}
\floatstyle{plaintop}
\restylefloat{table}
\usepackage{pgfplots}
\usepackage{times}
\ifCLASSINFOpdf
\else
\fi
\begin{document}
%
\title{Language Detection Engine for Multilingual Texting on Mobile Devices}


%
\author{\IEEEauthorblockN{Sourabh Vasant Gothe, Sourav Ghosh, Sharmila Mani, Guggilla Bhanodai, Ankur Agarwal, Chandramouli Sanchi}
\IEEEauthorblockA{Samsung R\&D Institute Bangalore, Karnataka, India 560037\\}
Email: \{sourab.gothe, sourav.ghosh, sharmila.m, g.bhanodai, ankur.a, cm.sanchi\}@samsung.com }


\makeatletter
\def\ps@IEEEtitlepagestyle{
  \def\@oddfoot{\mycopyrightnotice}
  \def\@evenfoot{}
}
\def\mycopyrightnotice{
  {\footnotesize
  \begin{minipage}{\textwidth}
  \centering
  \hrule\vspace{6pt}
  978-1-7281-6332-1/20/\$31.00~\copyright~2020 IEEE.
  Personal use of this material is permitted.  Permission from IEEE must be obtained for all other uses, in any current or future media, including reprinting/republishing this material for advertising or promotional purposes, creating new collective works, for resale or redistribution to servers or lists, or reuse of any copyrighted component of this work in other works.\\
  https://doi.org/10.1109/ICSC.2020.00057
  \end{minipage}
  }
}

\maketitle

\begin{abstract}
 More than 2 billion mobile users worldwide type in multiple languages in the soft keyboard. On a monolingual
 keyboard, 38\% of falsely auto-corrected words are valid in another language. This can be easily avoided by detecting the language
 of typed words and then validating it in its respective language. Language detection is a well-known problem in natural language processing. In
 this paper, we present a fast, light-weight and accurate Language
 Detection Engine (LDE) for multilingual typing that dynamically
 adapts to user intended language in real-time. We propose a novel
 approach where the fusion of character \textit{N}-gram model \cite{vatanen2010language} and
 logistic regression \cite{yu2011dual} based selector model is used to identify the
 language. Additionally, we present a unique method of reducing
 the inference time significantly by parameter reduction technique.
We also discuss various optimizations fabricated across LDE to
resolve ambiguity in  input text among the languages with the
same character pattern.
Our method demonstrates an average accuracy of 94.5\% for
Indian languages in Latin script and that of 98\% for European
languages on the code-switched data. This model outperforms
fastText \cite{joulin2016bag} by 60.39\%  and ML-Kit\footnote{https://firebase.google.com/docs/ml-kit/android/identify-languages} by 23.67\% in F1 score \cite{goutte2005probabilistic} for European languages. LDE is faster
on mobile device with an average inference time of 25.91$\mu$ seconds.

\end{abstract}




%

\section{\textbf{Introduction}}
In the current era of social media, language detection is a much required intelligence in mobile device for many applications viz. translation, transliteration, recommendations, etc.  Language detection algorithms work almost accurately when the language scripts are distinct using simple script detection method. In India, there are 22 official languages, and almost every language has its own script but in general user prefers to type in Latin script.
As per our statistical analysis, 39.78\% of words typed in QWERTY layout are from Indian languages. Hindi is a popular Indian language, 22.8\% of Hindi language users use QWERTY keyboard for typing, that implies the need of support for languages written in Latin script.


Standard languages written in Latin script i.e typed in QWERTY keyboard are referred to as Macaronic languages. These languages can have the same character pattern with other languages, unlike standard ones. For example, when Hindi language is written in Latin script (Hinglish), the word ``somwar'' which means Monday, shares the same text pattern with English word ``Ran\textbf{\underline{somwar}}e'', in such cases, character-based probabilistic models alone fail to identify the exact language as probability will be higher for multiple languages. Also, the user may type based on phonetic sound of the word that leads to variations like,  ``somwaar'', ``somvar'', ``somvaar'' etc. which are completely user dependent.

The soft keyboard provides next-word predictions, word completions, auto-correction, etc. while typing. Language Models (LMs) responsible for those are built using Long Short-Term Memory Recurrent Neural Networks (LSTM RNN)
\cite{hochreiter1997long} based Deep Neural networks (DNN) model \cite{mikolov2011extensions} with a character-aware CNN embedding \cite{kim2016character}. We use knowledge distilling method proposed by Hinton et al. \cite{hinton2015distilling} to train the LM \cite{chen2015strategies}. 
Along with LMs, adding another DNN based model for detecting the language that executes on every character typed, will increase the inference time and memory and leads to lag in mobile device. Additionally, in soft-keyboards extensibility is a major concern. Adding one or more language in the keyboard based on the locality or discontinuing the support of a language should be effortless. 

Considering the above constraints into account, we present the Language Detection Engine (LDE), an amalgamation of character $N$-gram models and a logistic regression based selector model. The engine is fast in inferencing on mobile device, light-weight in model size and accurate for both code-switched (switching between the languages) and monolingual text. This paper discusses various optimizations performed to increase engine accuracy compared to DNN based solutions in ambiguous cases of the code-switched input text.

We also discuss how LDE performs on five Indian Macaronic languages Hinglish (Hindi in English), Marathinglish (Marathi in English), Tenglish (Telugu in English), Tanglish (Tamil in English), Benglish (Bengali in English) and four European languages- Spanish, French, Italian, and German. 
Since the typing layout is of English language (Latin script),
we term English as a primary language and all other languages as a secondary language.

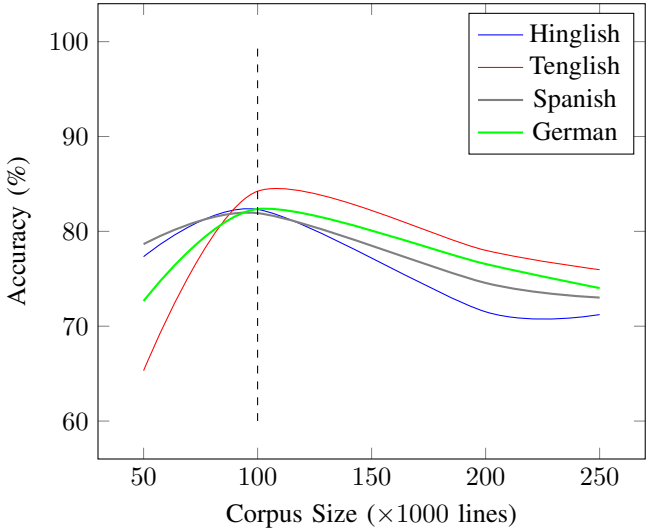
\begin{figure}
	\centering
	\pgfplotsset{
		width=\linewidth
	}
	\begin{tikzpicture}
	\begin{axis}[xlabel=Corpus Size ($\times 1000$ lines),ylabel=Accuracy (\%),
	xtick = {50, 100, 150, 200, 250},
	xticklabel style={/pgf/number format/fixed}
	]
	\addplot[smooth,blue] coordinates {
		(50,77.33)
		(100, 82.31)
		(200, 71.52)
		(250, 71.23)
	};
	\addlegendentry{Hinglish}
	\addplot[smooth,red,thin] coordinates {
		(50, 65.31)
		(100, 84.23)
		(200, 78.00)
		(250, 75.96)
	};
	\addlegendentry{Tenglish}
	\addplot[smooth,gray, thick] coordinates {
		(50, 78.65)
		(100, 81.92)
		(200, 74.56)
		(250, 73.02)
	};
	\addlegendentry{Spanish}
	\addplot[smooth,green, thick] coordinates {
		(50, 72.65)
		(100, 82.35)
		(200, 76.56)
		(250, 74.02)
	};
	\addlegendentry{German}
	\addplot[dashed] coordinates{
		(100, 60.00) (100, 100)
	};
	\end{axis}
	\end{tikzpicture}
	\caption{Char $N$-gram accuracy over corpus size}
	\label{fig:Corpus}
\end{figure}
 

\section{\textbf{Related Work}}
In this section, we discuss about work related to language detection, both $N$-gram based and deep learning based approaches.

\subsection{$N$-gram based models}
Ahmed et al. \cite{ahmed2004language} detail about language identification using $N$-gram based cumulative frequency addition to increase the classification speed of a document. It is achieved by reducing counting and sorting operations by cumulative frequency addition method. In our problem, we detect the language, based on user typed text rather than document with large information. Vatanen et al. \cite{vatanen2010language} compared the naive bayes classifer based charater $N$-gram and ranking method for language detection task. Their paper focuses on detecting short segments of length 5 to 21 characters, and all the language models are constructed independently of each other without considering the final classification in the process. We have adopted similar methodology for building char $N$-gram  models in our approach.

 Erik Tromp et al. \cite{tromp2011graph} discuss Graph-Based $N$-gram Language Identification (LIGA) for short and ill-written texts. LIGA outperforms other $N$-gram based models by capturing the elements of language grammar as a graph. However LIGA does not handle the case of code-switched text.
 
 All the above referred models do not prioritize recently typed words. For seamless multilingual texting which involves continuous code-switching, more priority must be given to the recent words so that the suggestions from currently detected language model can be fetched from the soft keyboard and shown to the user.

\begin{figure*}[t!]
	\centering
	\includegraphics[width=\textwidth]{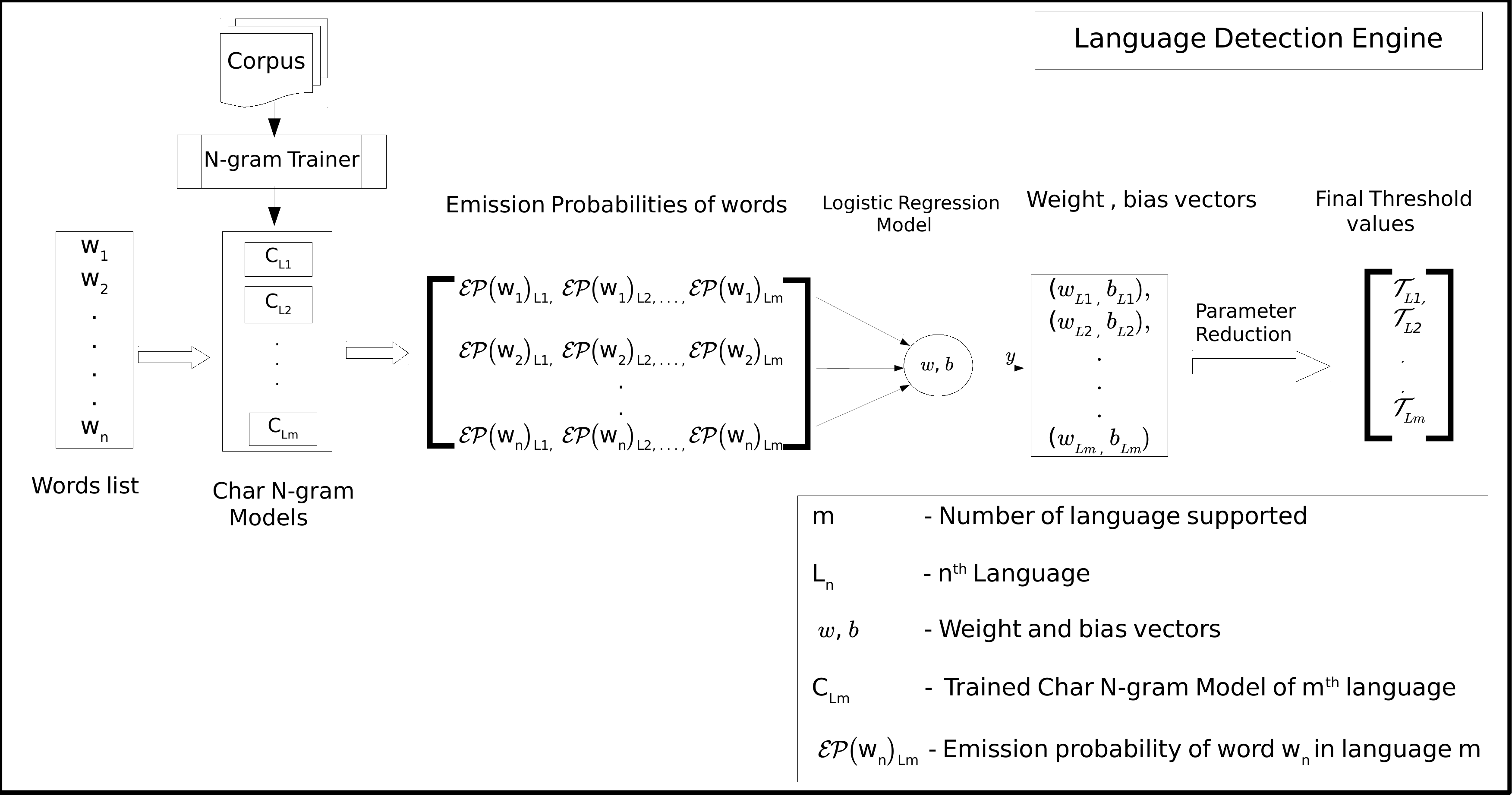}
	\caption{Model Training}
	\label{fig:ModelTraining}
\end{figure*}

\subsection{Deep learning based models}
Lopez et al. propose a DNN based language detection for spoken utterance \cite{lopez2014automatic} motivated by the success in acoustic modeling using DNN. They train a fully connected feed-forward neural network along with the logisic regression calibration to identify the exact language. Javier Gonzalez-Dominguez et al. \cite{gonzalez2014automatic} further extend to utilize LSTM RNNs for the same task. 

Zhang et al. \cite{zhang2018fast} have recently presented CMX, a fast, compact model for detection of code-mixed data. They address same problem as ours but in a different environment. They train basic feed-forward network that predicts the language for every token passed where multiple features are used to obtain the accurate results. However, such models require huge training data and are not feasible in terms of extensibility and model size for mobile devices.

This paper presents a novel method to resolve the ambiguity in input text and  detect the language accurately in multilingual soft-keyboard for five Indian macaronic languages and four European languges.

\section{\textbf{Proposed Method}}
We propose the Language Detection Engine (LDE) that enhances user experience in multilingual typing by accurately deducing the language of input text in real-time. LDE is a union of (a) character $N$-gram model which gives emission probability of input text originating from a particular language, and (b) a selector model which uses the emission probabilities to identify the most probable language for a given text using logistic regression \cite{fan2008liblinear}. Unique architecture of independent character $N$-gram models with selector model is able to detect the code-mixed multilingual context accurately.

\subsection{\textbf{Emission probability estimation using Character $N$-gram}}
Character $N$-gram is a statistical model which estimates probability distribution over the character set of a particular language given its corpus.
\subsubsection{Train data} 
Training corpus is generated by crawling online data from various websites. For Indian macaronic languages, we crawled the native script (Example: Devanagari script for Hindi) data of the languages and reverse transliterated to Latin script. This data is validated by the language experts for quality purpose.

We experimented with various sizes of corpus to train character $N$-gram model and found out that $100k$ sentences show the best accuracy from the model on sample test set, as detailed in Fig. \ref{fig:Corpus}.

\subsubsection{Model training} For every supported language $l_{i}$, we train Character $N$-gram model $C_{l_{i}}$ independent of the other languages as shown in Fig. \ref{fig:ModelTraining}. Probability of a sequence of words $\left(t_{1 .. n}\right)$ in language $l_{i}$ is given by,
\begin{equation}
\label{eqn:WordProbability}
P_{l_{i}}\left(t_{1 .. n}\right) = \prod_{k=1}^{n}\left.  P_{l_{i}}(t_k)^{r^{n-k}}\right.  ,\text{where }r \in \left(0, 1\right]
\end{equation}
We prioritize the probability of most recent word over the previous words using a variable $r$, value ranging between $0$ and $1$, that effectively reduces the impact of the leading words probability by converging values closer to $1$. To prevent the underflow of values we use logarithmic probabilities. Mathematically,
\begin{equation}
\label{eq:EmissionProbability}
\log{P_{l_{i}}\left(t_{1 .. n}\right)} = \sum_{k=1}^{n}\left.  r^{n-k}\cdot \log{P_{l_{i}}(t_k)}\right.
\end{equation}

The probability of sequence of characters in a word $t$, represented as $c_{0..m}$, where $c_{0}$ is considered as space character, is given by
\begin{equation}
\label{eqn:CharProbability}
P_{l_{i}}\left(c_{0 .. m}\right) = P_{l_{i}}\left(c_1 \middle| c_0\right) \cdot \prod_{k=2}^{m}\left. P_{l_{i}}\left(c_k \middle| c_{k-2}c_{k-1}\right)\right.
\end{equation}

These trained models $C_{l_{i}}$ are used to estimate the emission probability of character sequence during the inference for language $l_{i}$. We have chosen $n$ to be 3 in $N$-gram model, i.e character tri-gram model is trained on the corpus.

\subsection{\textbf{Selector Model}}
Here we briefly discuss the motivation behind an additional selector model. Firstly, input text originating from one language may also have significant emission probability in another language that may belong to the same family. This is because of words sharing the similar roots and frequent usage of loan words. 

For example, the Spanish word \textit{``vocabulario''}  shares linguistic root with its English counterpart \textit{``vocabulary''}. Again, \textit{``jungle''} which is a frequently used word in English, is actually a loan word via Hindi from Sanskrit. Presence of such words in the training corpus increases the preplexity of the model, i.e emission probabilities will be higher for multiple languages which makes it difficult to deduce the ultimate language.

Secondly, as character $N$-gram model gets trained based on the frequency of characters in the corpus, this makes it dependent on the size of character set. So, the emission probability values become incomparable as languages with smaller character set will statistically get higher values. Hence, these probabilities from character $N$-gram are not sufficient enough to determine the source language accurately. To this end, we present a logistic regression based selector model which addresses the illustrated problems.

Selector model $S$ comprises of weight and bias vectors $w$ and $b$ respectively of size $m$, where $m$ is the number of supported languages. This model transforms the emission probability provided by char $N$-gram such that the new probability value of word $t_{n}$, $P'_{l_{i}}\left(t_{n}\right)$, given by,
\begin{equation}
\label{eqn:LR}
\log P'_{l_{i}}\left(t_{n}\right) = w_{l_{i}}\cdot\log P_{l_{i}}\left(t_{ n}\right) + b_{l_{i}}
\end{equation}
where $l_{i}$ is deemed to be the origin language of word $t_{ n}$ if,
\begin{equation}
\label{eqn:LR_comp}
P'_{l_{i}}\left(t_{n}\right) \geq 0.5
\end{equation}
\subsubsection{Train data}
Training data for the selector model is
the vector of  emission probabilities of word $t_{n}$ for every language $l_{i}$. Batch of 200$k$ labeled words are used for training parameters of a particular language $l_{i}$. These 200$k$ words comprise of 100$k$ vocabulary words belonging to language $l_{i}$, and another 100$k$ words that are equally distributed among other languages.

Trained character $N$-gram models $C_{l_{1..m}}$ provide the required emission probabilities for every word from $m$ different languages.
 
\subsubsection{Model Training}
Weight and bias vectors of the selector model are trained such that for every input word, probability for labeled language is greater than $0.5$ as given in equation (\ref{eqn:LR_comp}). These trained weight and bias vectors are used to obtain new probability values as given in equation (\ref{eqn:LR}), which now become comparable among languages. Newly estimated probabilities resolve the ambiguity in the input text among the languages which have same patterns, with clearly dominated probability for the final detected language.

 As shown in Fig. \ref{fig:ModelTraining}, selector model takes emission probabilities $\varepsilon p (w_{n})_{L_{m}}$ for every word $w_{n}$ from each language ($L_{m}$) as input from pre-trained character $N$-gram models ($C_{l_{m}}$) and yields weight and bias vectors 
$w_{l_{i}}, b_{l_{i}}$ respectively.

\subsubsection{Parameter Reduction}
LDE performs a set of computations to detect the language on mobile device, which we term as on-device inference.
For every character typed by the user on a soft-keyboard, on-device inferencing happens followed by the inference of DNN Language model to provide next word predictions, word completions, and auto-correction, etc. based on the context.

To optimize on-device inference time, we propose a novel method of parameter reduction which reduces multiple computations during inference to a single arithmetic operation. Equation (4) is simplified to combine the weight and bias parameters as a single threshold value $\tau_{l}$ given by Equation (\ref{eq:threshold}) which effectively reduces the computation to constant time.

From Equations (\ref{eqn:LR}) and (\ref{eqn:LR_comp}), 
\begin{equation*}
\log P'_{l_{i}}\left(t_{n}\right) \geq \log 0.5
\end{equation*}
\begin{equation}
w_{l_{i}}\cdot\log P_{l_{i}}\left(t_{ n}\right) + b_{l_{i}} \geq \log 0.5
\label{eq:reduce}
\end{equation}
This can be further reduced to
\begin{equation*}
\begin{split}
	\log P_{l_{i}}\left(t_{n}\right) - \frac{\log 0.5 - b_{l_{i}}}{w_{l_{i}}}                                     & \geq 0        \\
	\implies   \log P_{l_{i}}\left(t_{ n}\right) - \frac{\log 0.5 - b_{l_{i}}}{w_{l_{i}}} + \log 0.5               & \geq \log 0.5 \\
	\implies   \log P_{l_{i}}\left(t_{ n}\right) - \frac{\left(w_{l_{i}} - 1\right)\cdot\log 2 - b_{l_{i}}}{w_{l_{i}}} & \geq \log 0.5
\end{split}
\end{equation*}
\begin{equation}
 \\
\therefore \log P_{l_{i}}\left(t_{ n}\right) - \tau_{l_{i}}                       \geq \log 0.5
\label{eq:finalInference}
\end{equation}
where $\tau_{l_{i}}$ is a parameter given by 
\begin{equation}
\label{eq:threshold}
\tau_{l_{i}} = \frac{\left(w_{l_{i}} - 1\right)\cdot\log 2 - b_{l_{i}}}{w_{l_{i}}}
\end{equation}
\subsubsection{On-device inference}
From equations (\ref{eq:reduce}) and (\ref{eq:finalInference}), it is evident that we can obtain the ultimate probability $\log P'_{l_{i}}\left(t_{n}\right)$ just by subtracting the threshold value $\tau_{l}$ from the logarithmic emission probability $\log P_{l}\left(t_{ n}\right)$ as given below,
\begin{equation}
\log P'_{l_{i}}\left(t_{n}\right) = \log P_{l_{i}}\left(t_{ n}\right) - \tau_{l_{i}}
\label{eq:final}
\end{equation}
where ${l_{i}}$ is the language and $t_{ n}$ is the character sequence. This makes probabilities from different languages comparable. 


\section{\textbf{Engine Architecture}}
\begin{figure}
	\centering
	\includegraphics[width=1\linewidth]{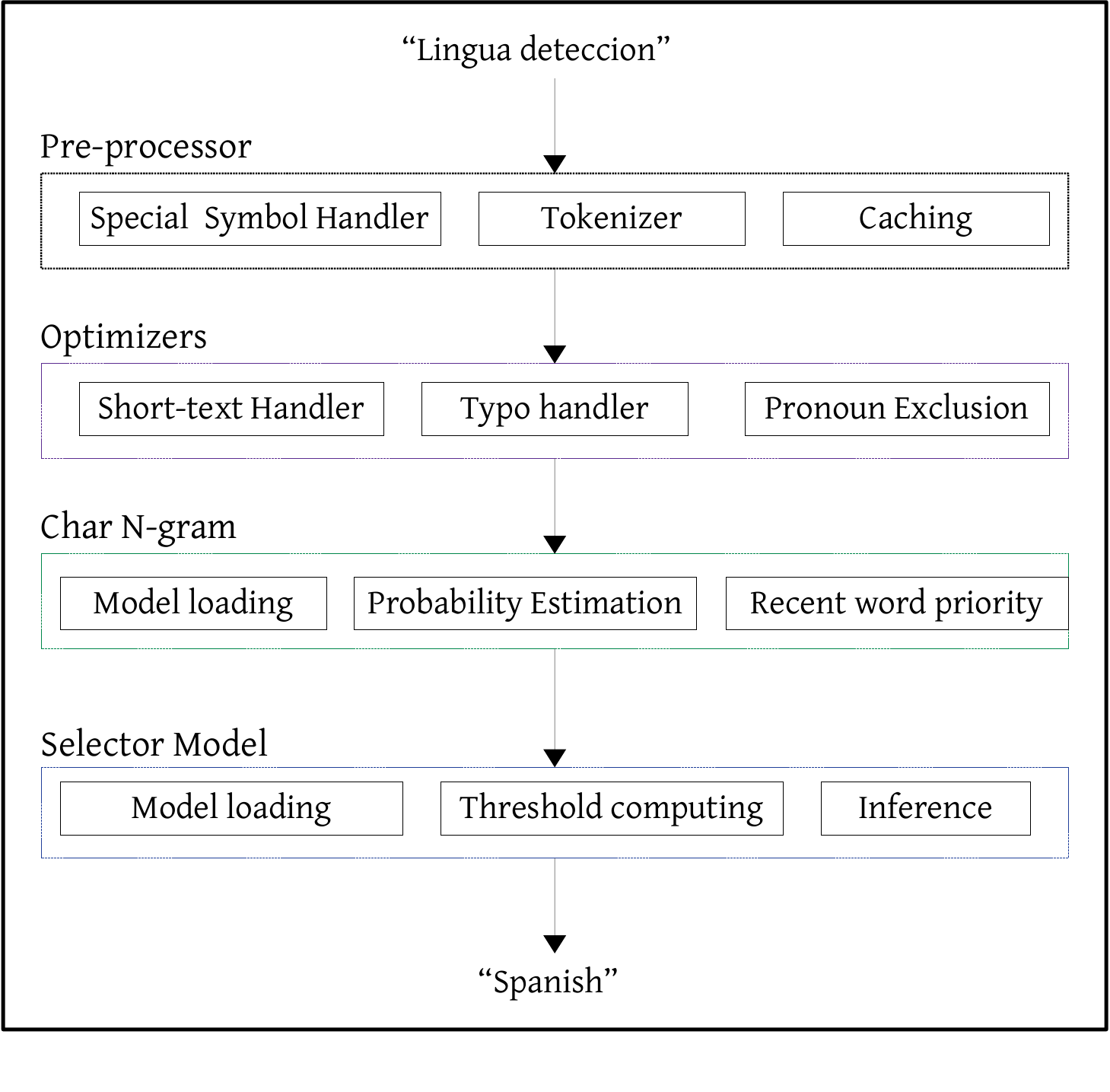}
	\caption{Engine Architecture}
	\label{fig:ModelArchitecture}
\end{figure}

Language Detection Engine constitutes of multiple components in various phases like preprocessor, optimizer, Char N-gram and Selector. The input text is first pre-processed and passed to the optimization phase where multiple heuristics are applied to address the enigmatic cases and proceeds to char $N$-gram inference and finally language selector phase to obtain the detected language. End-to-end architecture is represented in Figure \ref{fig:ModelArchitecture}. In this section, we explain each phase of the engine in detail.

\subsection{\textbf{Pre-processor}}
In this phase, the input text is preprocessed to obtain the required information from large context.

\subsubsection{Special Symbol Handler} 
In soft keyboard, the input may not be only text but can also include various ideograms like emojis, stickers, etc. 
This handler trims the input and provides the data that is necessary to detect the language.

\subsubsection{Tokenizer} Engine tokenizes the input context into tokens with whitespace as a delimiter. The last two tokens are concatenated and processed for language detection, which is observed to be most efficient in terms of processing time and accuracy, compared to considering more than two tokens. For short words with character length $\leq 2$, tokenizing is left to short-text handler.

\subsubsection{Caching} Based on the current detected language multiple algorithms like, auto-correction, auto-capitalization, touch-area correction \cite{azenkot2012touch} \cite{thomas2015hand} 
etc. tune the word suggestions accordingly in real-time. This leads to multiple calls to LDE for the same input text, hence LDE caches the language of previously typed text, to avoid the redundant task of detecting the language again. %

\subsection{\textbf{Optimizers}}
LDE addresses enigmatic cases in multilingual typing by applying additional optimizations that are discussed below.

\subsubsection{Short-text Handler} 
The context is an entire input that the user has typed and engine uses the previous two tokens of the context to detect the language. In the cases of short-words with character length less than or equal to two, it becomes ambiguous to detect the language. For example, ``to me'' is a valid context in English as well as in Hinglish, in such cases words before this context helps to deduce the exact source language. So extending the context to prior words, when context word length is less than two resolves the ambiguity for the engine to decide upon short words. We observed  ${{\sim}5\%}$ improvement in the accuracy of Indian macaronic languages with this change.

\subsubsection{Typo Handler} When a user makes a typo, often its harder to decide from which language the suggestions or correction should be provided. To address this issue, LDE obtains a correction candidate word from non-current language LM \cite{chen2015strategies} with an edit distance of one and effectively avoids the decrease in False Negatives due to wrong auto-correction. Below example illustrates the need for this heuristic,

``$\left[Hello\right]_{E} \left[bhai \; suno\right]_{H} \left[can \; we \; meet \; \right]_{E} [ksl]_{*}$'' 

where subscript $[]_{E}$ indicates English text and $[]_{H}$ as Hinglish and $[]_{*}$ a typo. When this context is typed in English and Hinglish bilingual keyboard, the engine fetches one auto-correction candidate $[kal]$ (meaning tomorrow) with an edit distance of one. Though the previous two words are from English, LDE manages to auto-correct the typo into valid word from non-current language Hinglish.

Typo handler automatically adopts to the user behavior while typing and provides valid corrections from the LM. For Indian macaronic languages, ${\sim}22\%$ improvement and for European languages ${\sim}15\%$ improvement observed in the F1 score of auto-correction on a linguist written bi-lingual test set.

In a closed beta trial with $2000$ soft keyboard users over a period of two months, 38\% of the falsely auto-corrected words are valid in another language. LDE is able to suppress the false auto-corrections and improve the auto-correction performance by 43.71\%\ in mono-lingual keyboard.
\subsubsection{Pronoun Exclusion} Practically, there is no particular language associated with proper nouns alone, but it follows the language of the entire context. To address this, the engine stores a linguist validated pronoun's list as a TRIE data-structure \cite{mani2019real} for efficient look-up. If the typed word is found in pronouns list, the cached language for the input excluding pronoun is considered as the detected language.

\begin{figure}
	\centering
	\includegraphics[width=1\linewidth]{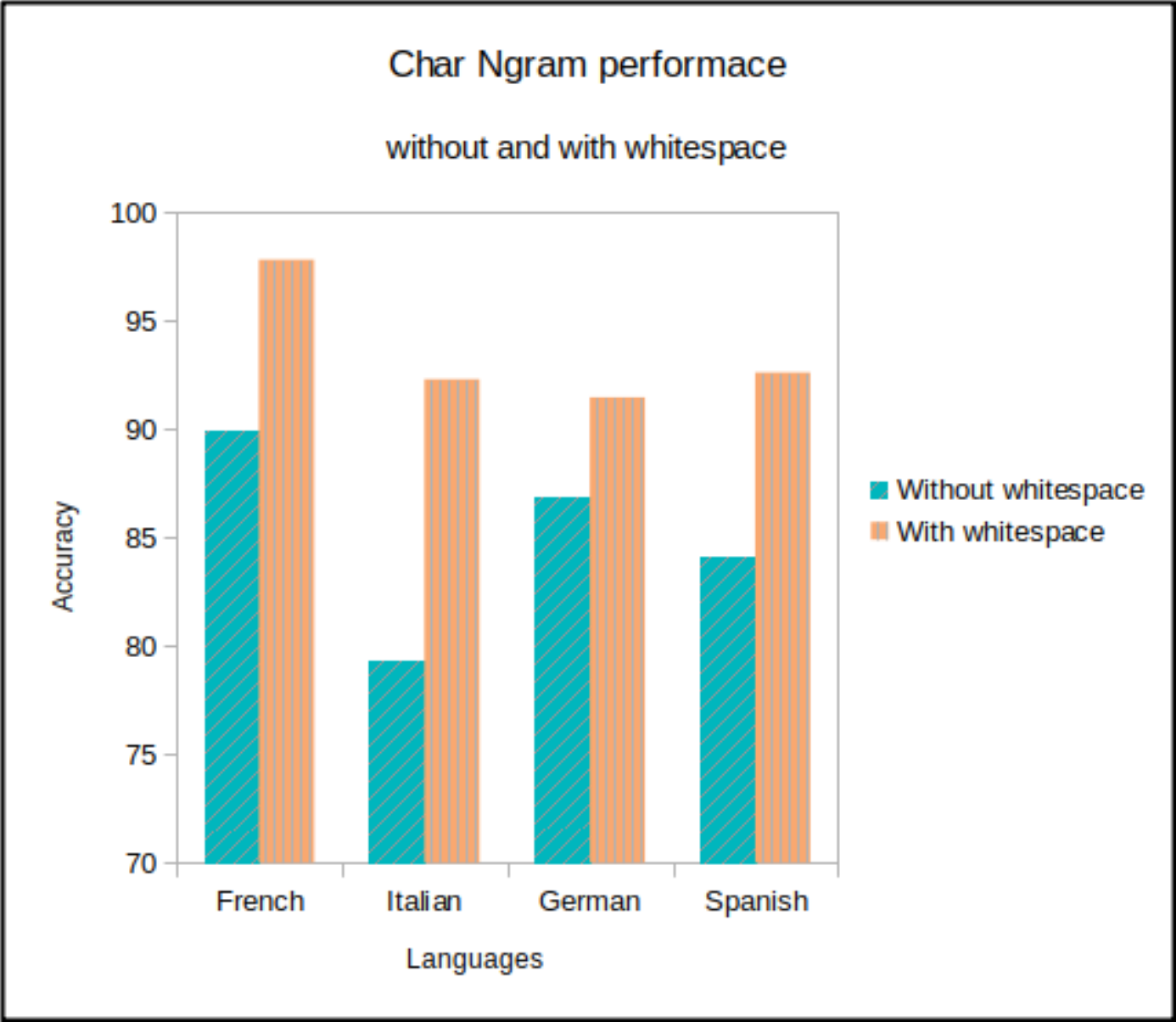}
	\caption{Pictorial representation of Table \ref{tab:InterTable}}
	\label{fig:chart0}
\end{figure}

\subsection{\textbf{Char $N$-gram}}
\subsubsection{Model loading} As explained in section 3. a tri-gram model is used to obtain emission probabilities of the character sequence. So there are $^{n+1}P_{3}$ possible character sequences in a language with the character set of length $n$ and an additional character, i.e whitespace ` '. These probabilities are pre-computed for every language and stored separately in a binary data file which is further compressed using zlib \cite{gailly2008zlib} compression to reduce the ROM size on mobile device.  Considering the whitespace \cite{ahmed2004language} as an extra character for training the char $N$-gram model makes an impact when the character pattern is same among multiple languages. Fig.  \ref{fig:chart0} depicts average gain of 10.13\% achieved for European languages when whitespace is considered.

For loading model on device, data file is uncompressed and probabilities are loaded to an array of every language.
Due to the modularity of model files, we can upgrade the model or add a new language and remove existing one just by training the required language's model. Such provision addresses the extensibility issue for soft-keyboard effectively.

\subsubsection{Recent word prioritization} In multilingual typing code-switching happens continuously, input begins with one language and eventually switches to another. To detect the current language in real-time, priority should be given to the recently typed character sequence as mentioned before.

In below example,

``$\left[Our \; company \;is  \;\right]_{E} \left[intentando\; \right]_{ES}$''

first three words are of English and the next is of Spanish. Ideally, current detected language should be Spanish as there is a code-switch. But if all characters are treated equally most probable language will be English. From Equation (\ref{eqn:WordProbability}) it can be observed that our char $N$-gram prioritizes trailing words than the leading words resulting in detecting the language accurately.
\subsection{\textbf{Selector Mode}l}
\subsubsection{Threshold computing} As explained in section III. A, logistic regression model is trained using library provided by sklearn \cite{fan2008liblinear}  in python. For every language, the model is trained to obtain the weight and bias, and further reduced to determine threshold value as explained in Equation (\ref{eq:threshold}). The complete processing is done on a 64-bit linux machine offline and threshold values are loaded to the model.

\subsubsection{Model loading}
After parameter reduction every individual language has corresponding threshold value. The threshold values are stored in  respective languages char $n$-gram data file itself and unload to an array of thresholds corresponding to supported languages on device. In this way, we curtail the effort of re-training all the models for any modifications and update only threshold values in respective data file. LDE does not require any large infrastructure to train and build the model, all our experiments were conducted on a linux machine of 4GB RAM.

\section{\textbf{Experimental Results}}
We compare the Language detection Engine performance with various baseline solutions like fastText library \cite{joulin2016bag},  langID.py \cite{Lui:2012:LOL:2390470.2390475} and Equilid a DNN model \cite{jurgens-etal-2017-incorporating} and also with Google's ML-Kit\footnote{https://developers.google.com/ml-kit}. In this section, we briefly explain the experimental set-up configured for all of above mentioned models and discuss about the test set that we prepared for the evaluation. Performance of LDE is compared with monolingual models like fastText and langId.py, ML-Kit and multilingual model such as Equilid.

\subsection{\textbf{fastText}}
Joulin et al. \cite{joulin2016bag} have distributed the model\footnote{https://dl.fbaipublicfiles.com/fasttext/supervised-models/lid.176.ftz} which can identify 176 languages. We used this model to compare the performance of European languages with LDE. However fastText pre-trained model does not support Indian macaronic languages.

\subsubsection*{\textbf{Custom fastText model for Indian macaronic languages}} 
We trained a custom fastText supervised model. We used reverse transliterated corpus of all the Indian languages which is validated by linguists. The same corpus is used to train LDE so that evaluation is comparable. 2.5GB of corpus was used to train the fastText model for five Indian languages each of size 500MB. Custom trained model size after quantization is 900KB.

\subsection{\textbf{ML-Kit}} ML Kit supports total of 103 languages including one Indian macaronic language, Hinglish. ML-Kit doesn't have a provision to train custom models for other Indian macaronic languages.
 For the experiment purpose a sample android application was developed that uses the API exposed by ML-Kit to identify the language and calculate the F1 score 
 for given test set. Complete evaluation was performed on Samsung Galaxy A50 device. 

\subsection{\textbf{Langid.py}} Langid.py is a standalone python tool by Lui and Baldwin \cite{Lui:2012:LOL:2390470.2390475} \cite{lui2011cross} that can identify 97 languages. Langid.py is monolingual model, i.e it can not identify code-switched text. Therefore we compare only on inter-sentential sentences where no code-switching is involved within a sentence. 

\subsection{\textbf{Equilid: Socially-Equitable Language Identification}} Jurgens et al.  propose a sequence-to-sequence DNN model \cite{jurgens-etal-2017-incorporating} for detecting the language. Equilid identifies the code-switched multilingual text and tags every word with the detected language. An experiment was conducted on a GPU  to evaluate the metric by loading the pre-trained models\footnote{http://cs.stanford.edu/~jurgens/data/70lang.tar.gz}. Pre-trained model is of size $559$MB which can identify 70 languages but none of the Indian macaronic languages are supported by Equilid.

\subsection{\textbf{Performance Evaluation}}
We evaluate the performance on two types of test sets based on code-switching style (a) Intra-sentential and (b) Inter-sentential. These test sets are hand written by the language experts involving natural code-switching. We evaluate above described methodologies and compare with LDE. 

\begin{table}[ht]
	\small
	\centering
	\resizebox{\columnwidth}{!}{%
		\begin{tabular}{c c c c c}
			\toprule
			{Language}   & Words    & Characters        & Code-switch (\%) &  \\ \midrule
			French     & $6430$ & $37968$ & $48.84$     &  \\
			Italian    & $4403$ & $30016$          & $53.17$    &  \\
			German     & $5499$ & $34380$ & $49.97$    &  \\
			Spanish    & $6663$ & $41231$          & $47.89$    &  \\ 
			Hinglish    & $6332$ & $36656$ & $61.70$     &  \\
			
			Benglish    & $6123$ & $34983$ & $59.25$         &  \\
			
			Marathinglish & $5520$ & $38580$ & $65.40$         &  \\
			
			Tanglish    & $6024$ & $35416$ & $57.29$         &  \\
			
			Tenglish    & $5958$ & $44676$ & $50.22$         &  \\ \bottomrule
			
		\end{tabular}%
	}
	\caption{Description of the Intra-sentential test set}
	\label{tab:TestSet1Desc}
\end{table}

\begin{table}[ht]
	\small
	\centering
	\resizebox{\columnwidth}{!}{%
		\begin{tabular}{c c c c c c}
			\toprule
			\multirow{4}{*}{Language} &                \multicolumn{4}{c}{F1 score}                 &  \\
			\cmidrule{2-6}       & fastText & LDE               & ML-Kit   & Equilid           &  \\ \midrule
			French           & $0.6714$ & $\textbf{0.9980}$ & $0.813$  & $0.9722$          &  \\
			Italian          & $0.7445$ & $0.9901$          & $0.7926$ & $\textbf{0.9934}$ &  \\
			German           & $0.5456$ & $\textbf{0.9960}$ & $0.8008$ & $0.9535$          &  \\
			Spanish          & $0.5144$ & $0.9870$          & $0.8044$ & $\textbf{0.9912}$ &  \\
			Hinglish          & $0.5232$ & $\textbf{0.9920}$ & $0.9120$  & $-$               &  \\
			Benglish          & $0.7562$ & $\textbf{0.9561}$ & $-$      & $-$               &  \\
			Marathinglish       & $0.6278$ & $\textbf{0.9840}$ & $-$      & $-$               &  \\
			Tanglish          & $0.7820$ & $\textbf{0.9765}$ & $-$      & $-$               &  \\
			Tenglish          & $0.7120$ & $\textbf{0.9981}$ & $-$      & $-$               &  \\ \bottomrule
		\end{tabular}%
	}
	\caption{Comparison on Intra-sentential test set}\label{tab:IntraTable}
\end{table}

\subsubsection{\textbf{Intra-sentential test set}}

In this type, the code-switching can occur anywhere in the sentence, where there are again two possibilities, i) Test set 1: context written mainly in primary language English and partly in secondary languages, for example,

\textit{``Can you believe midterms \textbf{\textit{comienza}} next week''}

where Spanish word is used while typing in English.

and ii) Test set 2: context written mainly in secondary language and partly written in primary language. For example,

\textit{``Justo \textbf{thinking} en ti''}

where English word is used while typing in Spanish.

A uniformly distributed test set of these two types were taken by picking $300$ sentences from each one. Every word in a test sentence is manually tagged with the source language. As there will be multiple code-switching involved, context level language detection is performed i.e based on previous two words the current language is identified which is exactly the way as LDE identifies the language for soft keyboard.

Statistics for these test sets like the percentage of code-switching involved, characters, words are shown in Table \ref{tab:TestSet1Desc}.

\begin{figure}
	\centering
	\includegraphics[width=1\linewidth]{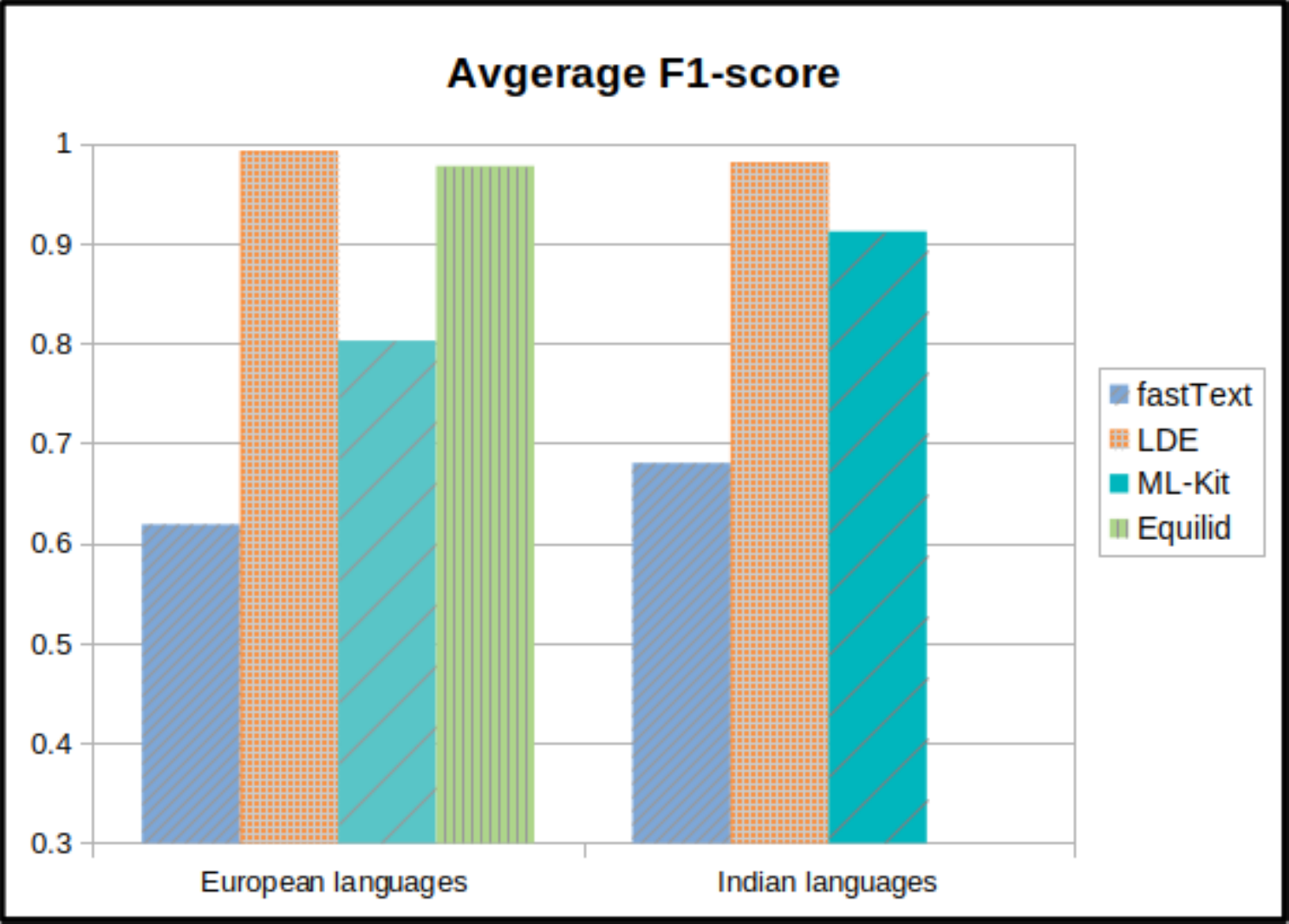}
	\caption{Pictorial representation of Table \ref{tab:IntraTable}}
	\label{fig:chart1}
\end{figure}

\subsubsection*{\textbf{F1 score}}
Table \ref{tab:IntraTable} shows the comparison of F1 score between fastText \cite{joulin2016bag}, ML-Kit, Equilid \cite {jurgens-etal-2017-incorporating} with LDE for European and Indian macaronic languages. For European languages LDE outperforms fastText  by 60.39\% and exceeds Google's ML-Kit by 23.67\% also surpasses Equilid DNN Model by 1.55\%. For Indian macaronic languages LDE is 44.29\% better than fastText and exceeds by 7.6\% for Hinglish with ML-Kit. It can be observed that LDE performs better than the DNN based models which are huge in model size.

Fig. (\ref{fig:chart1}) represents the visualization of the performance of various language detection models on intra-sentential test-set, where LDE is in par with the Equilid and significantly dominates fastText and ML-Kit.

\begin{table}[ht]
	\small
	\centering
	\resizebox{\columnwidth}{!}{%
		\begin{tabular}{c c c c c c}
			\toprule
			\multirow{4}{*}{Language} &                 \multicolumn{4}{c}{F1 score}                 &  \\
			\cmidrule{2-6}       & fastText & LDE               & ML-Kit            & LangId.py &  \\ \midrule
			French           & $0.9874$ & $0.9872$          & $\textbf{0.9962}$ & $0.9590$  &  \\
			Italian          & $0.9745$ & $0.9856$          & $0.9834$          & $0.8918$  &  \\
			German           & $0.9895$ & $\textbf{0.9901}$ & $0.9899$          & $0.902$   &  \\
			Spanish          & $0.9765$ & $0.9823$          & $\textbf{0.9892}$ & $0.9182$  &  \\
			Hinglish          & $0.9094$ & $\textbf{0.9530}$ & $0.912$           & $-$       &  \\
			Benglish          & $0.8563$ & $\textbf{0.9163}$ & $-$               & $-$       &  \\
			Marathinglish       & $0.6696$ & $\textbf{0.8936}$ & $-$               & $-$       &  \\
			Tanglish          & $0.7963$ & $\textbf{0.8696}$ & $-$               & $-$       &  \\
			Tenglish          & $0.8675$ & $\textbf{0.9102}$ & $-$               & $-$       &  \\ \bottomrule
		\end{tabular}%
	}
	\caption{Comparison on Inter-sentential test set}\label{tab:InterTable}
\end{table}
\subsubsection {\textbf{Inter-sentential test set}}
  In this type of data the code-switching occurs only after a sentence in first language is completely typed. Total of $500$ test sentences from every language combination are used to obtain the metric. Unlike in previous case, here we evaluate sentence level accuracy for each model, as there is no code-switching involved within the sentence. Additionally, we evaluated the same test set on LangID.py \cite{Lui:2012:LOL:2390470.2390475} which is a popular off-the-shelf model for this type of data.

\subsubsection*{\textbf{F1 score}}
\begin{figure}
	\centering
	\includegraphics[width=1\linewidth]{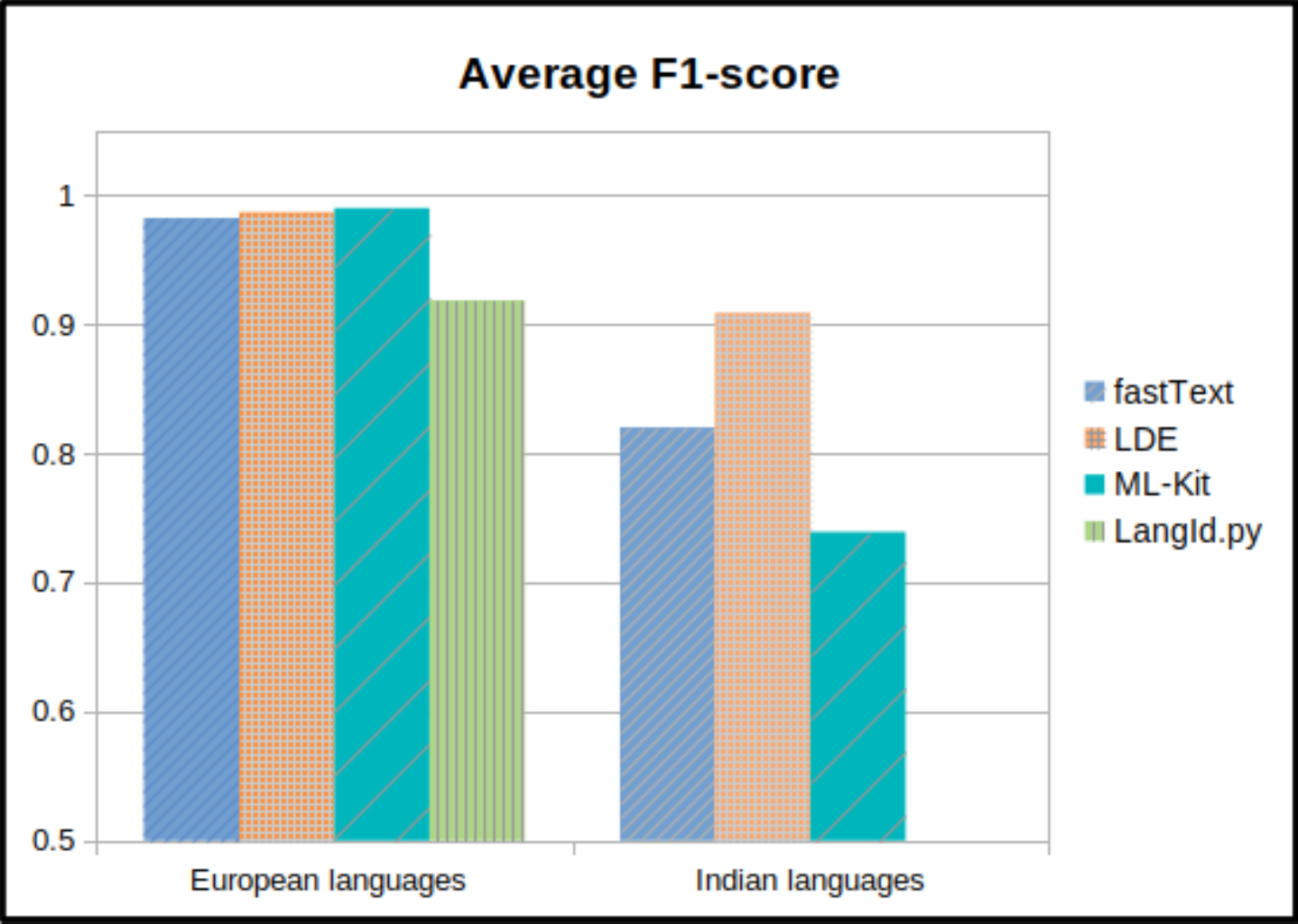}
	\caption{Pictorial representation of Table \ref{tab:InterTable}}
	\label{fig:chart2}
\end{figure}

 On inter-sentential test-set all the models perform accurately as there is a long context to identify. Table \ref{tab:InterTable} shows the F1 score for fastText, ML-Kit and LangId for European and Indian languages. It is observed that LDE is in par with ML-Kit and fastText and better than LangId.py for European languages. However, for Indian languages LDE dominates fastText by 10\% and ML-Kit by 22.95\% for Hinglish which shows that LDE performs as good as the DNN models. Figure \ref{fig:chart2}. shows the comparison of various models performance on inter-sentential test set.

\subsubsection*{\textbf{Inference time}}
Table \ref{tab:Inference time} shows the inference time and model size for all 10 supported languages on a uniformly distributed intra-sentential and inter-sentential test set. Average inference time is 25.91$\mu$ seconds and the model size of LDE for all $10$ languages combined is $166.65$KB.

\begin{table}[ht]
	\small
	\centering
	\resizebox{\columnwidth}{!}{%
		\begin{tabular}{c c c c}
			\toprule
			{Language}   & Inference Time ($\mu$s) & Model size (KB) &  \\ \midrule
			French     & $24.30$         & $22.23$    &  \\
			English     & $20.64$         & $20.28$    &  \\
			Italian    & $20.04$         & $21.71$    &  \\
			German     & $21.08$         & $18.44$    &  \\
			Spanish    & $24.30$         & $16.66$    &  \\
			Hinglish    & $35.41$         & $13.54$    &  \\
			Benglish    & $27.34$         & $13.46$    &  \\
			Marathinglish & $26.94$         & $14.31$    &  \\
			Tanglish    & $32.46$         & $13.74$    &  \\
			Tenglish    & $26.56$         & $12.28$    &  \\ \bottomrule
		\end{tabular}%
	}
	\caption{Average Inference time and Model size}
	\label{tab:Inference time}
\end{table}

\section{Conclusion}
We have proposed LDE a fast, light-weight, accurate engine for multilingual typing with a novel approach, that unites char $N$-gram and logistic regression model for improved accuracy.
LDE model size is 5X smaller than that of fastText custom trained model and $ {\sim}60\%$ better in accuracy. LDE  being a shallow learning model, either surpasses or in par with state-of-the-art DNN models in performance. Though char $N$-gram is trained on monolingual data, LDE accurately detects code-switching in a multilingual text with the help of uniquely designed selector model. LDE also improved the performance of auto-correction by 43.71\% by suppressing correction of valid foreign words. Furthermore, LDE is suitably designed for supporting extensibility of languages.






%
\bibliographystyle{IEEEtran}
\bibliography{LanguageDetectionEngine}\nocite{*}

\end{document}